\pdfoutput=1

\documentclass[11pt]{article}

\usepackage[final]{acl}

\usepackage{times}
\usepackage{latexsym}

\usepackage[T1]{fontenc}

\usepackage[utf8]{inputenc}

\usepackage{microtype}

\usepackage{inconsolata}

\usepackage{indentfirst}
\usepackage{CJKutf8}
\usepackage[normalem]{ulem}
\usepackage{tabularx}
\usepackage[most]{tcolorbox}
\usepackage{ragged2e}
\usepackage{multirow} 
\usepackage{multicol} 
\definecolor{my-blue}{cmyk}{0.80, 0.13, 0.14, 0.04, 1.00}
\usepackage[bb=dsserif]{mathalpha}
\usepackage{bbm}
\usepackage{lipsum}
\usepackage{booktabs}

\title{CharPoet: A Chinese Classical Poetry Generation System Based on Token-free LLM}

\author{Chengyue Yu$^{* \dag}$, Lei Zang$^{* \dag}$, Jiaotuan Wang, Chenyi Zhuang, Jinjie Gu
        \\ Ant Group
        \\ \{yuchengyue.ycy, zanglei.zl, yunting.wjt,chenyi.zcy, jinjie.gujj\}@antgroup.com
        }

\begin{document}
\begin{CJK*}{UTF8}{gbsn}
\maketitle
\begin{abstract}
Automatic Chinese classical poetry generation has attracted much research interest, but achieving effective control over format and content simultaneously remains challenging. Traditional systems usually accept keywords as user inputs, resulting in limited control over content. Large language models (LLMs) improve content control by allowing unrestricted user instructions, but the token-by-token generation process frequently makes format errors. Motivated by this, we propose CharPoet, a Chinese classical poetry generation system based on token-free LLM, which provides effective control over both format and content. Our token-free architecture generates in a character-by-character manner, enabling precise control over the number of characters. Pruned from existing token-based LLMs, CharPoet inherits their pretrained capabilities and can generate poetry following instructions like ``Write me a poem for my mother's birthday.'' CharPoet achieves format accuracy above 0.96, outperforming Jiuge-GPT-2 (0.91) and GPT-4 (0.38). In terms of content quality, CharPoet surpasses traditional systems including Jiuge, and is comparable to other LLMs. Our system is open source and available at \url{https://modelscope.cn/models/CharPoet/CharPoet}. A video demonstration of CharPoet is available at \url{https://youtu.be/voZ25qEp3Dc}.
\end{abstract}

\def\thefootnote{$\ast$}\footnotetext{Equal contribution.}
\def\thefootnote{$\dag$}\footnotetext{Corresponding authors.}

\begin{figure}
    \centering    \includegraphics[width=0.47\textwidth]{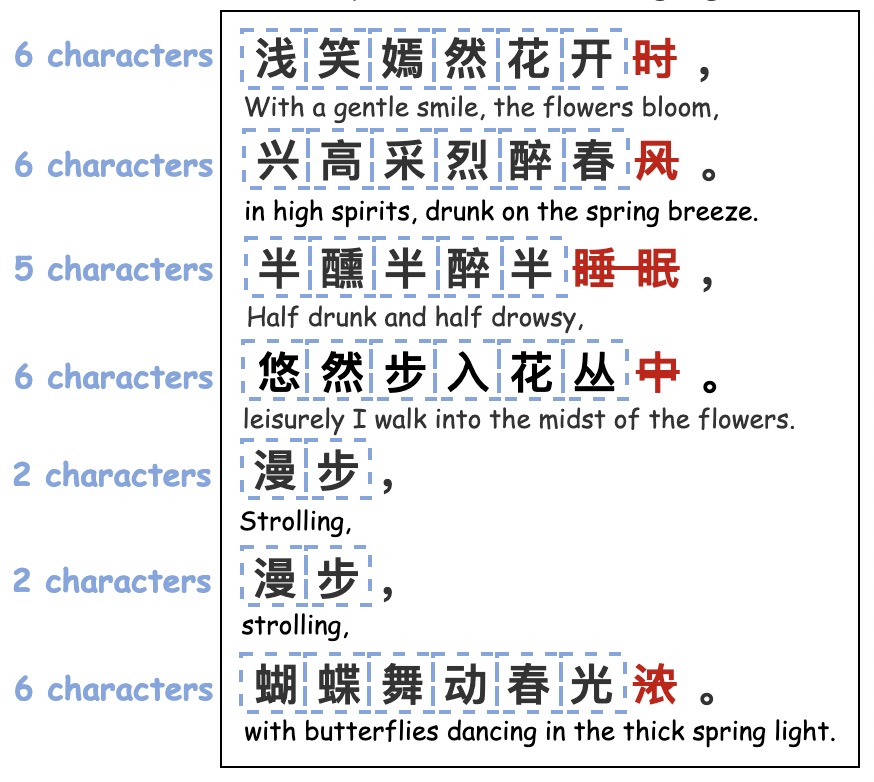}
    \raggedright
    \caption {Poem generated by GPT-4. The poem violates the format requirement of \textit{Rumengling} with 6 excess characters.}
    \label{fig:gpt4_case}
\end{figure}
\section{Introduction}

Chinese classical poetry, one of the most valuable heritages of human culture, conveys rich connotations through its concise and exquisite form. Chinese classical poetry can be classified into two primary categories: SHI and CI, both of which have strict format requirements \cite{hu2020generating}. For example, Wuyanjueju, the simplest form of SHI, requires four lines with each line containing exactly five Chinese characters. CI is more complex: there are nearly one thousand forms in total, each with different requirements for the number of lines and characters.

Automatic generation of Chinese classical poetry has attracted much research interest. However, achieving effective control over both format and content simultaneously remains a challenge.

Traditional systems in this field usually take keywords as user inputs \cite{guo2019jiuge,hu2020generating,wang2016chinese,yan2016poet,yi2017generating,yi2018chinese,zhang2014chinese,zhang2017flexible}. However, it is often insufficient for users to fully describe the theme or emotion they expect with just one or several keywords. This inability to process complex inputs has reduced the diversity and quality of the generated poetry. In contrast, Large Language Models (LLMs) can accept unrestricted user prompts and allow more control over the content. LLMs are capable of generating diversified texts following complex user instructions \cite{openai2022chatgpt,openai2023gpt4,bai2023qwen}. Nevertheless, token-based LLMs face challenges in strictly adhering to the expected format of poetry, occasionally producing lines with an excess or insufficient number of characters.

An example of a GPT-4-generated poem is given in Figure 1. In this example, GPT-4 is asked to write a poem in the \textit{Rumengling} form, with the keyword \textit{cheerful}. The generated poem performs well in terms of content, but it clearly violates the format requirements. The redundant characters are marked in red with a strikethrough.

We argue that the problem is partly due to the token-based nature of LLMs. Standard token-based LLM systems split text into word pieces before feeding them into the model. These text pieces are known as \textit{tokens}, and they usually contain more than one character \cite{sennrich2016neural, schuster2012japanese}. The system must generate text in a token-by-token manner. Under such a setting, if a model needs to control the number of characters precisely, it must know exactly how many characters are contained in each token. We have conducted a simple test that shows LLMs clearly lack such knowledge. The results are provided in Appendix A.

Motivated by this, we propose CharPoet, a Chinese classical poetry generation system based on a token-free LLM, which achieves effective control over both format and content simultaneously. ``Token-free'' here means that our model operates only on characters or bytes, in contrast to regular tokens. As shown in Figure 2, our system generates poems in a character-by-character manner. With the token-free architecture, our system can precisely control the number of characters. Instead of being trained from scratch, our token-free LLM is pruned from existing token-based models. We remove long tokens from the tokenizer and the language model head, keeping only character-level and byte-level tokens, and then finetune on a poetry dataset. Through this pruning process, our system inherits capabilities from existing token-based LLMs, and can generate poetry following complex instructions such as ``Write me a poem for my mother's birthday.''

Without any post-processing, our token-free system achieves a format accuracy of 0.96, outperforming Jiuge-GPT-2 (0.91) and GPT-4 (0.38). In addition, our system performs comparably to existing LLMs in terms of the content quality.

\begin{figure}
    \centering    \includegraphics[width=0.49\textwidth]{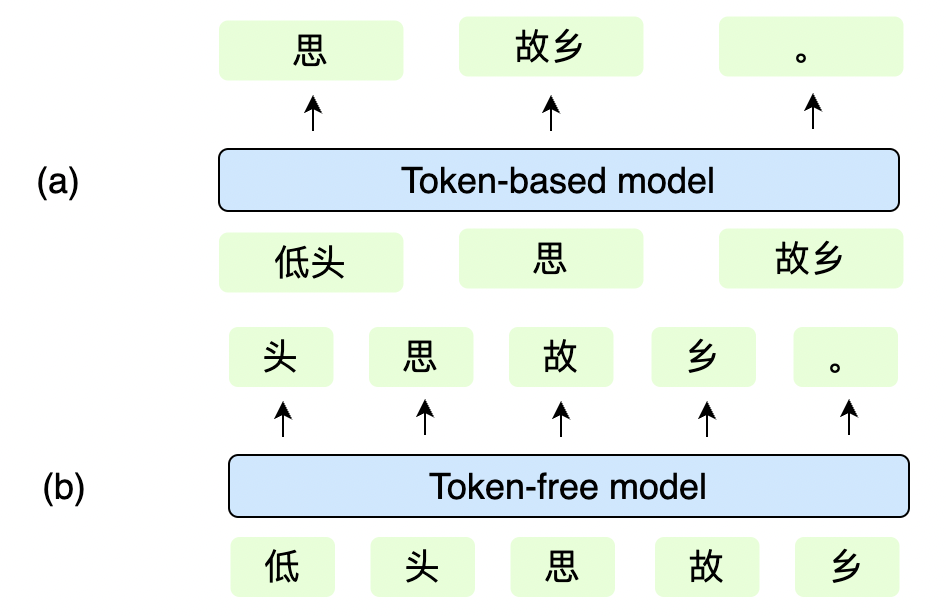}
    \raggedright
    \caption{Generation process of a token-based model vs. a token-free model: (a) In a token-based model, the system may output more than one character at a time, resulting in difficulty in exerting precise control over the number of characters. (b) In a token-free model, the system outputs at most one character at a time, making control over the number of characters easier.}
    \label{fig:generation}
\end{figure}

\section{Related work}

\begin{figure*}[h!]
    \centering    \includegraphics[width=1\textwidth]{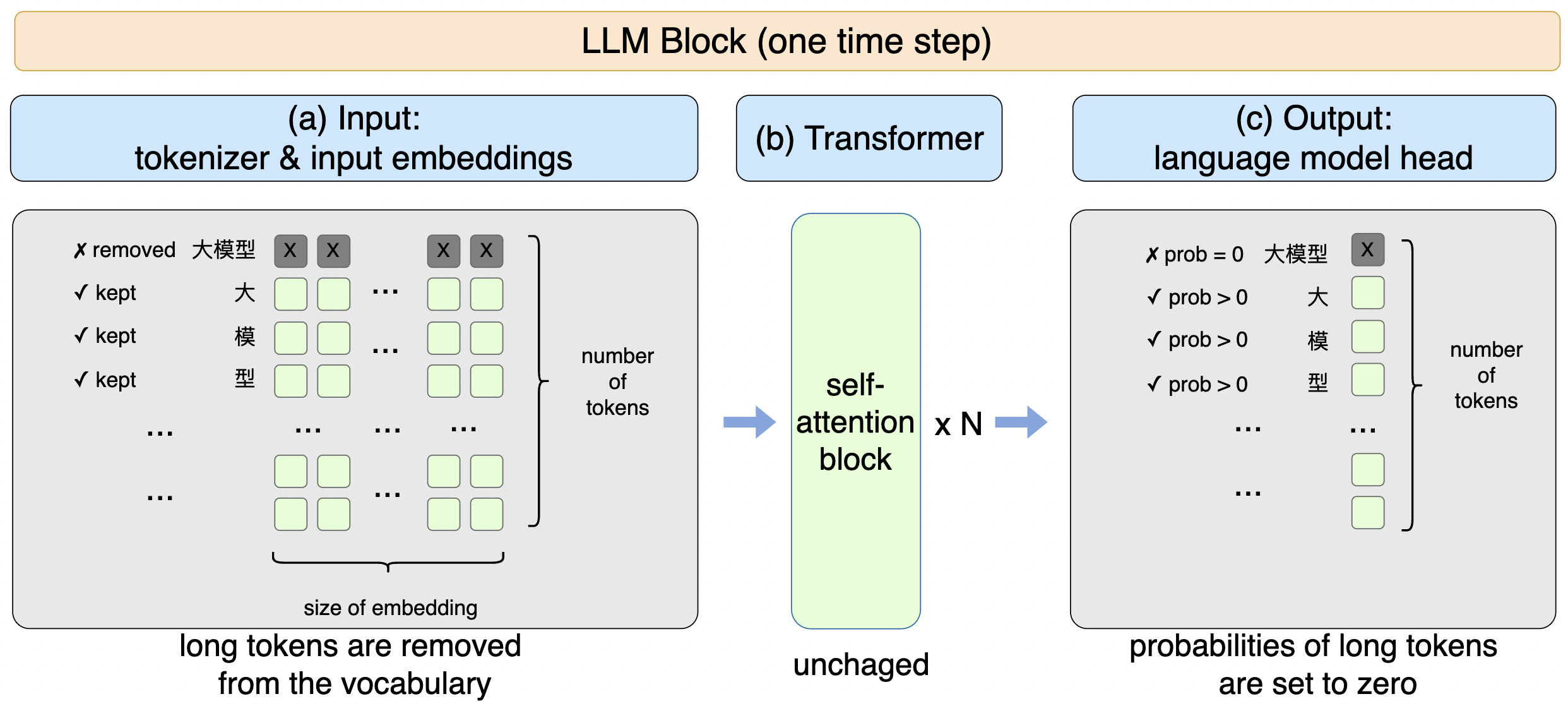}
    \raggedright
    \caption{Prune a token-based model into a token-free model. (a) For \textit{Input}, long tokens will be removed from the vocabulary. Text would only be tokenized into character-level or byte-level tokens; The embeddings of long tokens will never be accessed. (b) Transformer structure is left unchanged. (c) For \textit{Output}, the logits of long tokens will be set to a large negative number and the probabilities of long tokens will be zero. The language model head would never produce long tokens.}
    \label{fig:pruning}
\end{figure*}

\textbf{Traditional systems} in this field \cite{zhang2014chinese, wang2016chinese, yan2016poet, yi2017generating, yi2018chinese, guo2019jiuge} have demonstrated that RNNs and LSTMs can generate high-quality poetry. However, these systems usually accept keywords as user inputs, resulting in poor control over content. Moreover, they often have complex architectures or special modules designed to handle the strict format and content constraints inherent in poetry. For example, \newcite{yi2018chinese} imposes a working memory mechanism; \newcite{guo2019jiuge} implements a postprocess module to filter poems with unexpected format. 

\textbf{Large Language Models (LLMs)} \cite{openai2022chatgpt,openai2023gpt4,bai2023qwen} have demonstrated the power of the Transformer architecture \cite{vaswani2017attention} when trained with a large corpus. LLMs are capable of generating high-quality and diversified poetry following unrestricted prompt. However, they suffer from problems with format accuracy due to their token-based nature.

More in line with our research, \newcite{hu2020generating, belouadi2023bygpt5} build poetry generation systems based on \textbf{token-free language models}. However, those systems are trained from scratch so they do not inherit the great power from pretrained LLMs. They still accept keywords as user inputs and cannot understand complex instructions.

\section{Architecture}

\subsection{Pruning}

The core of our system is a token-free LLM. Instead of being trained from scratch as in previous work \cite{belouadi2023bygpt5}, our token-free LLM is pruned from an existing token-based LLM to inherit its pretrained knowledge and capabilities. Our token-free model accepts unrestricted prompts as input and returns poems that excel in both format accuracy and content quality.

We have designed a procedure that can prune any typical token-based LLM into a token-free model. A typical LLM, such as Llama \cite{touvron2023llama} and Qwen \cite{bai2023qwen}, contains three components, the \textit{Input} (the tokenizer and the input embeddings), the \textit{Transformer} \cite{vaswani2017attention} and the \textit{Output} (the language model head). Our pruning procedure modifies the \textit{Input} component and the \textit{Output} component, and leaves the \textit{Transformer} component unchanged. The procedure is described below and illustrated in Figure 3:

{\bfseries (a) Input Pruning.} We prune the tokenizer's vocabulary by removing all long tokens, leaving only character-level or byte-level fragments. \textit{Long tokens} refer to two types: tokens with more than one Chinese character and tokens consisting of a single Chinese character combined with non-Chinese characters. Once these tokens are removed, the tokenizer will only produce character-level or byte-level fragments. Subsequently, the input embeddings for these removed tokens will never be accessed or updated.

We retain non-Chinese tokens as they are. This approach ensures that the keywords commonly used in LLM chat settings like ``user'' and ``assistant'', remain intact to preserve the standard tokenization of chat templates.

{\bfseries (b) Transformer kept unchanged.} The structure of the Transformer is left unchanged, while the parameters will still be updated during finetuning.

{\bfseries (c) Output Pruning.} For outputs, we set the probabilities of all long tokens to zero. This is achieved by incorporating an indicator function into the original softmax transformation:
$$\text{Prob}(t_{i}) = \frac{ (1 - \mathbb 1_L (i) ) \exp(logit_i) }{\sum_{j} (1 - \mathbb 1_L (j) )\exp(logit_j)}
$$
where $logit_i$ denotes the neural network's output value of the \textit{i}th token prior to the softmax transformation, and the indicator function determines if the \textit{i}th token is a member of the long token set \textit{L}.
$$ \mathbb 1_L (x) = 
\begin{cases}
1\hspace{0.5cm} \text{if } x\in L\\
0\hspace{0.5cm} \text{if } x\notin L
\end{cases}
$$

In practice, we implement this by adding a large negative number to the logits of long tokens, instead of modifying the softmax function directly.

\begin{figure*}[h!]
    \centering
    \includegraphics[width=1\textwidth]{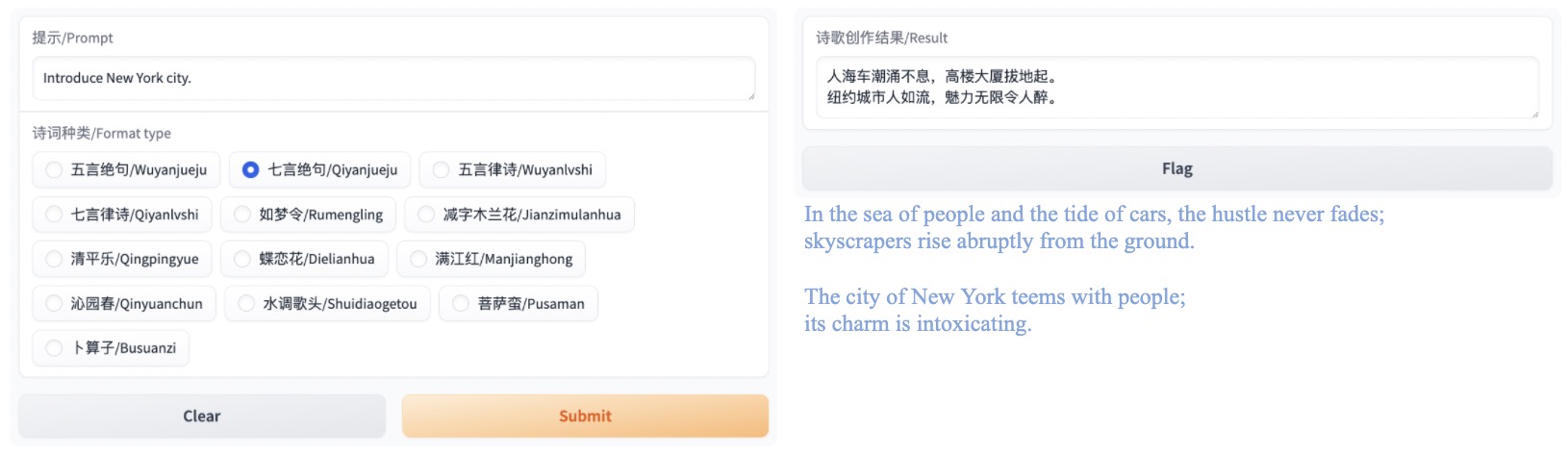}
    \caption{The user interface and generated poetry sample of CharPoet.}
    \label{fig:AI Poet}
\end{figure*}

With the above procedure, any typical LLM could be pruned to a token-free model. In contrast to typical token-based LLMs, the pruned token-free model outputs text in a character-by-character manner and is expected to perform better on character-sensitive tasks such as poetry generation. 

In this paper, we use Qwen-7B-Chat \cite{bai2023qwen} as the base model. An interesting observation is that even without further finetuning, the pruned token-free LLM is already capable of answering simple questions. We provide some examples in Appendix B. However, we suggest further finetuning on the target dataset in order to fully exploit the potential of the pretrained base model.

\begin{table*}[h]
\begin{tabularx}{0.99\textwidth}
{ 
  | >{\centering\arraybackslash}p{3.5cm} 
  | >{\centering\arraybackslash}X 
  | >{\centering\arraybackslash}p{2.1cm} 
  | >{\centering\arraybackslash}p{2.1cm}  
  | >{\centering\arraybackslash}p{2.1cm} 
  | >{\centering\arraybackslash}p{2.1cm} 
  |
}
\hline

\multirow{2}{*}{\bf Format Type} & \multirow{2}{*}{\bf \#Chars} &  {\bf GPT-4} & {\bf Jiuge-GPT-2} & {\bf Qwen (Finetuned)} & {\bf CharPoet (Ours)}  \\

\cline{3-6}
 & &  keyword / instruction & keyword / instruction & keyword / instruction & keyword / instruction  \\

\hline
\bf WuyanJueju (SHI)     & 20 &  0.49 / 0.73 & \bf 1.00 / - &     0.94 / \bf 1.00 &     0.98 /     0.99 \\
\bf WuyanLvshi (SHI)     & 40 &  0.29 / 0.36 & \bf 1.00 / - &     0.97 /     0.98 &     0.97 / \bf 0.99 \\
\bf QiyanJueju (SHI)     & 28 &  0.88 / 0.78 & \bf 1.00 / - &     0.99 / \bf 1.00 & \bf 1.00 / \bf 1.00 \\
\bf QiyanLvshi (SHI)     & 56 &  0.81 / 0.68 & \bf 1.00 / - &     0.98 /     0.96 &     0.97 / \bf 0.98 \\
\bf Rumengling (CI)     & 33 &  0.13 / 0.09 &     0.90 / - &     0.95 /    0.97 & \bf 1.00 / \bf 0.99 \\
\bf Jianzimulanhua (CI) & 44 &  0.81 / 0.79 &     0.96 / - &     0.99 /     0.97 & \bf 1.00 / \bf 0.99 \\
\bf Qingpingyue (CI)    & 46 &  0.13 / 0.18 &     0.96 / - & \bf 0.98 /     0.97 &     0.95 / \bf 0.99 \\
\bf Dielianhua (CI)     & 60 &  0.21 / 0.12 &     0.91 / - &     0.94 /     0.98 & \bf 0.99 / \bf 0.98 \\
\bf Manjianghong (CI)   & 93 &  0.07 / 0.04 &     0.83 / - &     0.88 /     0.90 & \bf 0.95 / \bf 0.95 \\
\bf Qinyuanchun (CI)    & 114 & 0.00 / 0.01 &     0.55 / - &     0.64 /     0.75 & \bf 0.82 / \bf 0.86 \\
\hline
\bf Avg         & 53.4 & 0.382 / 0.378 & 0.911 / - & 0.926 / 0.948 & \bf 0.963 / \bf 0.972 \\
\hline
\end{tabularx}
\caption{Evaluation on Format Accuracy. CharPoet outperforms other systems on average under both the keyword setting and the instruction setting. CharPoet performs significantly better than other systems on longer poems, such as \textit{Manjianghong} and \textit{Qinyuanchun}. }
\end{table*}

\subsection{Training}
Training involves two stages: general-purpose training and poetry-field training.
\subsubsection{General Purpose Training}
We need general-purpose training because our model is pruned from an existing token-based LLM. The original token-based model is trained on token sequences instead of character sequences, and is thus not familiar with natural language that is presented at character level. Here we use BELLE dataset \cite{ji2023better}, which is a high-quality general-purpose instruction-following dataset.

\subsubsection{Poetry-field Training}

In the second stage, we train with our in-house poetry dataset. The dataset contains 10000 SHI and 10000 CI, and each poem is associated with a human written \textit{prompt} related to it.

To improve format accuracy, we provide the model with a masked version of the expected poem as a format hint. In this \textit{masked poem}, all Chinese characters are replaced with a mask sign \textbf{[M]} while punctuation and line breaks are kept as is. An example of a \textit{masked poem} in the form \textit{Rumengling} is provided in below. \\

\begin{tcolorbox}[skin=widget,
boxrule=1mm,
coltitle=black,
colframe=my-blue!45!white,
colback=my-blue!15!white,
width=(.9\linewidth),
before=\hfill,
after=\hfill,
center]
\centering
\raggedright
\text{[M][M][M][M][M][M]，}\\
\text{[M][M][M][M][M][M]。}\\
\text{[M][M][M][M][M]，}\\
\text{[M][M][M][M][M][M]。}\\
\text{[M][M]，}\\
\text{[M][M]，}\\
\text{[M][M][M][M][M][M]。}
\end{tcolorbox}

The \textit{masked poem} is provided together with the original user prompt. The final prompt-response format is designed as follows, where \textbf{[SOP]} denotes \textit{start of piece}, \textbf{[EOP]} denotes \textit{end of piece}, \textbf{\{ORIGINAL USER PROMPT\}} denotes the original prompt from the user, \textbf{\{MASKED POEM\}} is the format hint, and \textbf{\{POEM\}} denotes the generated poem. \\

\begin{tcolorbox}[skin=widget,
boxrule=1mm,
coltitle=black,
colframe=my-blue!45!white,
colback=my-blue!15!white,
width=(.9\linewidth),
before=\hfill,
after=\hfill,
center]
\centering
\raggedright
\text{[SOP]}user\\
Fill in all the masks \text{[M]}.\\
\{ORIGINAL USER PROMPT\}\\
Output: \{MASKED POEM\}\\
\text{[EOP]}\\
\text{[SOP]}assistant\\
\{POEM\}\\
\text{[EOP]}\\
\end{tcolorbox}

In this way, the poem-generating task is transformed into a mask-filling task. With the token-free architecture, our model fills in all the masks in a character-by-character manner. The mask-filling design ensures that the model can strictly follow the format constraints of the requested poetry type.

\section{Demonstration}

The user interface of our poetry generation system is shown in Figure 4. In contrast to previous systems where users need to summarize the theme of the poetry they want in one or several keywords, our system allows users to describe desired content with natural language in the prompt box. After that, the user selects a poetry form and clicks the “Submit” button. A few seconds later, the system returns a poem following the user's instruction.

Our system is fully open source, available at \url{https://modelscope.cn/models/CharPoet/CharPoet}. We have included a Jupyter notebook in the project. Using this notebook, anyone can launch the application and try our system. We also provide some example poems in Appendix C.



\section{Evaluation}

\subsection{Test settings}
We evaluate performance on two aspects: format accuracy and content quality. Following \newcite{hu2020generating}, we assess performance on four types of SHI and six types of CI. We conduct tests under two user input settings: the first is the conventional keyword setting, where the user input consists of a single keyword; the second is the instruction setting, where the user input is a natural language instruction, such as ``Write me a poem for my mother's birthday.''. In both settings, one specific format is selected as the expected format.

We test 100 times for each type of form and each setting. For the keyword setting, we use 100 frequently used Chinese idioms collected from the internet. Chinese idioms convey rich meanings in simple expressions, and are thus more challenging than regular words. For the instruction setting, we ask GPT-4 to generate 100 prompts to eliminate potential human selection bias.

To exploit GPT-4's potential in format accuracy, we have carefully designed the prompt. We find that GPT performs better if provided with an example poem of the required form. The prompt template is provided in Appendix D.

\subsection{Models for Comparison}

We compare our system CharPoet with two categories of public available systems. One category is general-purpose LLMs, with GPT-4 \cite{openai2023gpt4} being the top performer. The other category is systems exclusively designed for automatic poetry generation, with Jiuge \cite{guo2019jiuge} being the most representative. 

Jiuge \cite{guo2019jiuge} is a comprehensive system with a postprocessing module to ensure format accuracy; therefore, when evaluating format accuracy, we compare instead with Jiuge-GPT-2 \cite{hu2020generating}, the most recent work in the Jiuge series, which is more comparable since it is transformer-based and end-to-end.

To verify the effectiveness of our token-free architecture, we also conducted an ablation study, where we compared our system to its token-based equivalent. The token-based equivalent is identical to CharPoet in every aspect including model size, prompt design and training dataset, except that it is built on the original token-based Qwen-Chat \cite{bai2023qwen} instead of our pruned token-free version. The token-based equivalent is marked as \textit{Qwen (Finetuned)} in later tables and figures.

\subsection{Evaluation on Format Accuracy}

Format accuracy results are shown in Table 1. A poem is counted as accurate only if the number of characters for every line is correct (perfect match). The figures for Jiuge-GPT-2 are directly collected from the original paper, while the figures for other models are obtained from our testing procedure. The figures for Jiuge-GPT-2 under the instruction setting are not available since Jiuge-GPT-2 does not support user instructions.

CharPoet performs better in format accuracy than all competing models, achieving an overall accuracy above 0.96 under both settings. Our ablation study comparing CharPoet with Qwen (Finetuned) confirms that the token-free architecture is effective, bringing a 3\% gain in format accuracy.

\begin{figure}
    \centering    \includegraphics[width=0.45\textwidth]{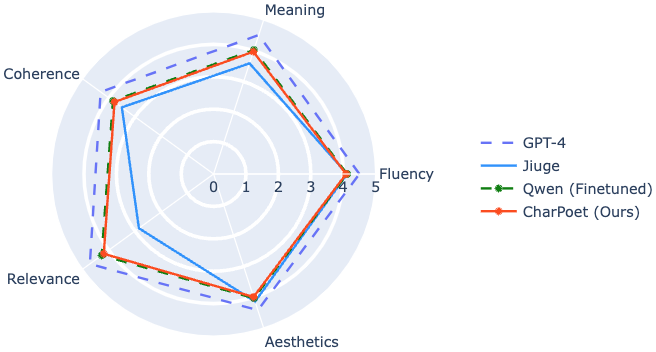}
    \raggedright
    \caption {Evaluation on Content Quality under the Keyword Setting.}
    \label{fig:radar_simple}
\end{figure}
\begin{figure}
    \centering    \includegraphics[width=0.44\textwidth]{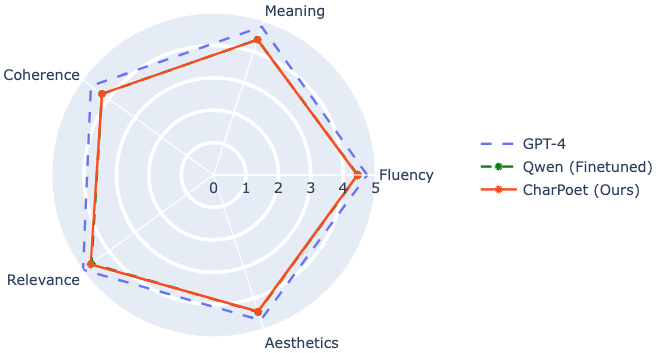}
    \raggedright
    \caption {Evaluation on Content Quality under the Instruction Setting.}
    \label{fig:radar}
\end{figure}

Consistent with \newcite{hu2020generating}, our results show that SHI is simple and all models listed here achieve decent accuracy. As for CI, which is more complex and challenging, our system beats previous systems by a large margin. For example, in terms of \textit{Qinyuanchun}, the longest type of poem in our test set, our system achieves 0.84 accuracy, compared to 0.55 of Jiuge-GPT-2 and nearly zero of GPT-4. Regression analysis also indicates that CharPoet is less sensitive to poem length (See Appendix E for details).

\subsection{Evaluation on Content Quality} 

Following \newcite{yi2018chinese}, we evaluate content quality with five criteria; each criterion needs to be scored on a 5-point scale:

    {\bfseries Fluency}. Does the poem obey the grammatical, structural and phonological rules?
    
    {\bfseries Meaning}. Does the poem convey some certain messages?
    
    {\bfseries Coherence}. Is the poem as a whole coherent in meaning and theme?
    
    {\bfseries Relevance}. Does the poem express user topics well?
    
    {\bfseries Aesthetics}. Does the poem have some poetic and artistic beauties?

Again to avoid bias, GPT-4 instead of human evaluators carries out the whole scoring process. The results under the two settings are shown in Figures 5 and 6. The performance of CharPoet is basically the same as that of Qwen (Finetuned) and not far from GPT-4, while it significantly surpasses Jiuge, especially in terms of \textit{Relevance}. The gain in content relevance indicates that pretrained LLMs can provide significantly better control over content compared to traditional models.

\section{Conclusion}

In this paper, we address the problem of achieving effective control over both format and content in the field of automatic Chinese classical poetry generation. We propose a token-free system CharPoet, which generates in a character-by-character manner, enabling precise control over the number of characters. Moreover, CharPoet allows for human instructions in natural language, in contrast to traditional models that only accept keywords.

CharPoet achieves format accuracy above 0.96 without any postprocessing, higher than Jiuge-GPT-2 (0.91) and GPT-4 (0.38). Our ablation study comparing CharPoet with its token-based equivalent shows that the token-free architecture brings a 3\% gain in format accuracy. In addition, our system's performance in content quality surpasses traditional systems, and is comparable to existing LLMs.


\newpage
\bibliography{acl_latex}

\begin{thebibliography}{20}
\expandafter\ifx\csname natexlab\endcsname\relax\def\natexlab#1{#1}\fi

\bibitem[{Belouadi and Eger(2023)}]{belouadi2023bygpt5}
Jonas Belouadi and Steffen Eger. 2023.
\newblock Bygpt5: End-to-end style-conditioned poetry generation with token-free language models.
\newblock In \emph{61st Annual Meeting of the Association for Computational Linguistics}. Association for Computational Linguistics (ACL).

\bibitem[{Guo et~al.(2019)Guo, Yi, Sun, Li, Yang, Liang, Chen, Zhang, and Li}]{guo2019jiuge}
Zhipeng Guo, Xiaoyuan Yi, Maosong Sun, Wenhao Li, Cheng Yang, Jiannan Liang, Huimin Chen, Yuhui Zhang, and Ruoyu Li. 2019.
\newblock Jiuge: A human-machine collaborative chinese classical poetry generation system.
\newblock In \emph{Proceedings of the 57th annual meeting of the association for computational linguistics: system demonstrations}, pages 25--30.

\bibitem[{Hu and Sun(2020)}]{hu2020generating}
Jinyi Hu and Maosong Sun. 2020.
\newblock Generating major types of chinese classical poetry in a uniformed framework.
\newblock In \emph{Proceedings of the Twelfth Language Resources and Evaluation Conference}, pages 4658--4663.

\bibitem[{Itzhak and Levy(2022)}]{itzhak2022models}
Itay Itzhak and Omer Levy. 2022.
\newblock Models in a spelling bee: Language models implicitly learn the character composition of tokens.
\newblock In \emph{Proceedings of the 2022 Conference of the North American Chapter of the Association for Computational Linguistics: Human Language Technologies}, pages 5061--5068.

\bibitem[{Ji et~al.(2023)Ji, Gong, Deng, Peng, Niu, Ma, and Li}]{ji2023better}
Yunjie Ji, Yan Gong, Yong Deng, Yiping Peng, Qiang Niu, Baochang Ma, and Xiangang Li. 2023.
\newblock \href {http://arxiv.org/abs/2304.07854} {Towards better instruction following language models for chinese: Investigating the impact of training data and evaluation}.

\bibitem[{Kaushal and Mahowald(2022)}]{kaushal2022tokens}
Ayush Kaushal and Kyle Mahowald. 2022.
\newblock What do tokens know about their characters and how do they know it?
\newblock In \emph{Proceedings of the 2022 Conference of the North American Chapter of the Association for Computational Linguistics: Human Language Technologies}, pages 2487--2507.

\bibitem[{OpenAI(2022)}]{openai2022chatgpt}
OpenAI. 2022.
\newblock \href {https://openai.com/blog/chatgpt} {Introducing {ChatGPT}}.

\bibitem[{OpenAI(2023)}]{openai2023gpt4}
OpenAI. 2023.
\newblock \href {http://arxiv.org/abs/2303.08774} {Gpt-4 technical report}.

\bibitem[{Ouyang et~al.(2022)Ouyang, Wu, Jiang, Almeida, Wainwright, Mishkin, Zhang, Agarwal, Slama, Ray, Schulman, Hilton, Kelton, Miller, Simens, Askell, Welinder, Christiano, Leike, and Lowe}]{ouyang2022training}
Long Ouyang, Jeff Wu, Xu~Jiang, Diogo Almeida, Carroll~L. Wainwright, Pamela Mishkin, Chong Zhang, Sandhini Agarwal, Katarina Slama, Alex Ray, John Schulman, Jacob Hilton, Fraser Kelton, Luke Miller, Maddie Simens, Amanda Askell, Peter Welinder, Paul Christiano, Jan Leike, and Ryan Lowe. 2022.
\newblock \href {http://arxiv.org/abs/2203.02155} {Training language models to follow instructions with human feedback}.

\bibitem[{Schuster and Nakajima(2012)}]{schuster2012japanese}
Mike Schuster and Kaisuke Nakajima. 2012.
\newblock Japanese and korean voice search.
\newblock In \emph{International Conference on Acoustics, Speech and Signal Processing}, pages 5149--5152.

\bibitem[{Sennrich et~al.(2016)Sennrich, Haddow, and Birch}]{sennrich2016neural}
Rico Sennrich, Barry Haddow, and Alexandra Birch. 2016.
\newblock Neural machine translation of rare words with subword units.
\newblock In \emph{54th Annual Meeting of the Association for Computational Linguistics}, pages 1715--1725. Association for Computational Linguistics (ACL).

\bibitem[{the Llama~team(2023)}]{touvron2023llama}
the Llama~team. 2023.
\newblock \href {http://arxiv.org/abs/2307.09288} {Llama 2: Open foundation and fine-tuned chat models}.

\bibitem[{the Qwen~team(2023)}]{bai2023qwen}
the Qwen~team. 2023.
\newblock \href {http://arxiv.org/abs/2309.16609} {Qwen technical report}.

\bibitem[{Vaswani et~al.(2017)Vaswani, Shazeer, Parmar, Uszkoreit, Jones, Gomez, Kaiser, and Polosukhin}]{vaswani2017attention}
Ashish Vaswani, Noam Shazeer, Niki Parmar, Jakob Uszkoreit, Llion Jones, Aidan~N Gomez, {\L}ukasz Kaiser, and Illia Polosukhin. 2017.
\newblock Attention is all you need.
\newblock \emph{Advances in neural information processing systems}, 30.

\bibitem[{Wang et~al.(2016)Wang, He, Wu, Wu, Li, Wang, and Chen}]{wang2016chinese}
Daisy~Zhe Wang, Wei He, Hua Wu, Haiyang Wu, Wei Li, Haifeng Wang, and Enhong Chen. 2016.
\newblock Chinese poetry generation with planning based neural network.
\newblock In \emph{Proceedings of COLING 2016, the 26th International Conference on Computational Linguistics: Technical Papers}, pages 1051--1060.

\bibitem[{Yan(2016)}]{yan2016poet}
Rui Yan. 2016.
\newblock i, poet: automatic poetry composition through recurrent neural networks with iterative polishing schema.
\newblock In \emph{Proceedings of the Twenty-Fifth International Joint Conference on Artificial Intelligence}, pages 2238--2244.

\bibitem[{Yi et~al.(2017)Yi, Li, and Sun}]{yi2017generating}
Xiaoyuan Yi, Ruoyu Li, and Maosong Sun. 2017.
\newblock Generating chinese classical poems with rnn encoder-decoder.
\newblock In \emph{Chinese Computational Linguistics and Natural Language Processing Based on Naturally Annotated Big Data: 16th China National Conference, CCL 2017, and 5th International Symposium, NLP-NABD 2017, Nanjing, China, October 13-15, 2017, Proceedings 16}, pages 211--223. Springer.

\bibitem[{Yi et~al.(2018)Yi, Sun, Li, and Yang}]{yi2018chinese}
Xiaoyuan Yi, Maosong Sun, Ruoyu Li, and Zonghan Yang. 2018.
\newblock Chinese poetry generation with a working memory model.
\newblock In \emph{Proceedings of the 27th International Joint Conference on Artificial Intelligence}, pages 4553--4559.

\bibitem[{Zhang et~al.(2017)Zhang, Feng, Wang, Wang, Abel, Zhang, and Zhang}]{zhang2017flexible}
Jiyuan Zhang, Yang Feng, Dong Wang, Yang Wang, Andrew Abel, Shiyue Zhang, and Andi Zhang. 2017.
\newblock Flexible and creative chinese poetry generation using neural memory.
\newblock In \emph{Proceedings of the 55th Annual Meeting of the Association for Computational Linguistics (Volume 1: Long Papers)}, pages 1364--1373.

\bibitem[{Zhang and Lapata(2014)}]{zhang2014chinese}
Xingxing Zhang and Mirella Lapata. 2014.
\newblock Chinese poetry generation with recurrent neural networks.
\newblock In \emph{Proceedings of the 2014 Conference on Empirical Methods in Natural Language Processing (EMNLP)}, pages 670--680.

\end{thebibliography}

\newpage

\appendix

\section{Probing into LLM's knowledge in token-character relationship}

For a token-based LLM, if it needs to control the number of characters precisely, it must know exactly how many characters are contained in each token. We have conducted a simple test, which shows that LLMs clearly lack such knowledge.

\subsection{Method}

Following \newcite{itzhak2022models}, we use a probing procedure called ``spelling bee'' to investigate how much a LLM knows about the token-character relationship of its vocabulary. Specifically, we probe whether the model has the knowledge that the token ``大模型'' contains three characters ``大'', ``模'' and ``型''.

The models we investigate here are the Qwen-series \cite{bai2023qwen}, including Qwen-1.7B-chat, Qwen-7B-chat and Qwen-14B-chat. The Qwen series is one of the earliest open-source LLMs with a strong ability in Chinese and is influential in the Chinese community. In the context of large language models, the probing procedure could be formulated as an instruction following task, designed as follows.\\

\begin{tcolorbox}[skin=widget,
boxrule=1mm,
coltitle=black,
colframe=my-blue!45!white,
colback=my-blue!15!white,
width=(.9\linewidth),
before=\hfill,
after=\hfill,
center,
adjusted title={
\centering
\textbf{Prompt} \\
\raggedright
List all the characters in the following token: <|extra\_1|>大模型
}
]
\centering
\textbf{Response}\\
\raggedright
大<|extra\_1|>模<|extra\_1|>型
\end{tcolorbox}\\

Here the special symbol <|extra\_1|> is used to ensure that both the long token in the prompt and the single characters in the response are tokenized as they are. We randomly selected 1000 tokens from the vocabulary to serve as a test set, and the remaining tokens are used as training examples.

Our procedure is not exactly the same as previous work \cite{kaushal2022tokens,itzhak2022models}. The main differences are
\begin{enumerate}
    \item Our experiment probes all language model parameters, while previous work \cite{kaushal2022tokens,itzhak2022models} probes only the vocabulary embedding, which ignores the knowledge contained in later layers, and thus would underestimate the real knowledge level.
    \item In the context of large language models, we are able to conduct probing experiments relatively easily with the model itself through supervised finetuning \cite{ouyang2022training}, while previous work probes with a separate model, which may be difficult to train and would also underestimate the real knowledge level.
\end{enumerate}

To sum up, we believe that our procedure can better estimate how LLM knows about the token-character relationship relationship.

\subsection{Results}

The results of the spelling bee probing procedure are summarized with the \textit{overall failure rate}, which is defined as the number of times that the LLM fails to correctly output the character sequence of the required token, divided by the size of the test set (which is 1000 in our experiment). We also pay attention to another version of the failure rate, where we count an output as a failure only if the number of characters in the output does not equal the length of the original token. This metric is more relevant since it directly influences the control of character numbers in poem generation. \\

\begin{table}[h]
\begin{tabularx}{0.47\textwidth}
{ 
  | >{\centering\arraybackslash}X
  | >{\centering\arraybackslash}p{2.1cm} 
  | >{\centering\arraybackslash}p{2.1cm}  
  |
}
\hline

\multirow{2}{*}{\bf Model Size} &  {\bf Token-based} & {\bf Token-free} \\

 &  overall / \#chars & overall / \#chars \\

\hline
\bf 1.8B &  15.1\% / 2.5\%  &   0.7\% / 0.2\% \\
\bf 7B   &   9.9\% / 1.7\%  &   0.5\% / 0.1\% \\
\bf 14B  &   4.1\% / 0.9\%  &   0.4\% / 0.0\% \\
\hline
\end{tabularx}
\caption{Failure rate of the spelling bee probing procedure. The token-free model has reduced the failure rate by more than 90\% compared to the token-based model, whether it's the overall failure rate or the failure rate by number of characters. }
\end{table}

We have several observations: 
\begin{enumerate}
    \item Failure rates of token-based models are non-trivial. Failure rate near 1\% seems small, but the influence is significant since mistakes would accumulate as the poem gets long. 
    \item The failure rate of token-based models seems not to vanish totally as model size increases.
    \item The token-free model has reduced the failure rate by more than 90\%.
\end{enumerate}

\newpage

\section{Token-free LLM without further finetuning}
Without further finetuning the pruned token-free LLM is already capable of answering simple questions. Examples are shown below.\\

\begin{tcolorbox}[skin=widget,
boxrule=1mm,
coltitle=black,
colframe=my-blue!45!white,
colback=my-blue!15!white,
width=(.9\linewidth),
before=\hfill,
after=\hfill,
center,
adjusted title={
\centering
\textbf{User:} \\
\raggedright
简要介绍一下日心说
}]
\centering
\textbf{Agent:} \\
\raggedright
日心说，也称日心地动说，是16th century末期由哥白尼提出的天体学说，它将日、月、水星、金星、木星、土星、天王星、海王星和冥王星等9大天体，及小天体如彗星、小星体等，都绕着以日为圆心的椭圆轨转。
\end{tcolorbox}\\

\begin{tcolorbox}[skin=widget,
boxrule=1mm,
coltitle=black,
colframe=my-blue!45!white,
colback=my-blue!15!white,
width=(.9\linewidth),
before=\hfill,
after=\hfill,
center,
adjusted title={
\centering
\textbf{User:} \\
\raggedright
Briefly introduce the heliocentric theory. 
}]
\centering
\textbf{Agent:} \\
\raggedright
The heliocentric theory is a model of the solar system that places the Sun at the center, with the planets orbiting around it in椭ical paths. This theory was proposed by Nicolaus Copernicus in the 16th century, but it was not widely accepted until the work of later astronomers such as Johannes Kepler and Galileo Galilei. The heliocentric model is now the most widely accepted model of the solar system, and it is supported by a vast amount of observational evidence.
\end{tcolorbox}\\

We can see that the main meaning is correct, although there are some typos and a mix of Chinese and English.

\newpage

\section{Example Poems Generated by CharPoet}

In the first example, our system is asked to write a poem with the keyword \textit{cheerful} and the required form is \textit{Rumengling}. This is exactly how we test GPT-4 in the Introduction section. We can see that poem generated by our system precisely match the required format of \textit{Rumengling}, in contrast to GPT-4. \\

\begin{tcolorbox}[skin=widget,
boxrule=1mm,
coltitle=black,
colframe=my-blue!45!white,
colback=my-blue!15!white,
width=(.9\linewidth),
before=\hfill,
after=\hfill,
center]

\centering
笑口频开深院，更说秋风天气。\\
With a frequent smile in the deep courtyard, I praise the autumn weather.\\
心事向人知，却好兴高采烈。\\
I share thoughts with others, in high spirits and delight.\\
休觅，休觅，酒到不知醒地。\\
Seek no more, seek no more. Drink till you do not know where you are.\\
\end{tcolorbox}\\

More importantly, CharPoet allows for unrestricted instructions. In the following example, the prompt is ``Write me a poem for my mother's birthday.'' and the required form is \textit{Rumengling}. The output follows the expected format and content.\\

\begin{tcolorbox}[skin=widget,
boxrule=1mm,
coltitle=black,
colframe=my-blue!45!white,
colback=my-blue!15!white,
width=(.9\linewidth),
before=\hfill,
after=\hfill,
center]
\centering
\raggedright

生日恰逢今日，母爱万金难拟。\\
Today is your birthday. A mother's love is priceless. \\
恩重更情浓，岁岁同歌同醉。\\
Your kindness is profound. We sing and celebrate together, in each passing year. \\
同醉，同醉，寿星高上天际。\\
Drink together, drink together. You are the star today.
\end{tcolorbox}

\newpage
\section{Test the Performance of GPT-4 on Format Accuracy}

To exploit GPT-4's potential in format accuracy, we have carefully designed the prompt. We find that GPT performs better if we provide it with an example poem of the required form. Our prompt is designed as follows. \\

\begin{tcolorbox}[skin=widget,
boxrule=1mm,
coltitle=black,
colframe=my-blue!45!white,
colback=my-blue!15!white,
width=(.9\linewidth),
before=\hfill,
after=\hfill,
center,
]
\centering
\textbf{Prompt} \\
\raggedright
请写一首如梦令，主题或要求为“兴高采烈”。请严格按照如梦令对每一句话的字数要求，下面给出一个例子：\\
\centering
常记溪亭日暮，沉醉不知归路。\\
兴尽晚回舟，误入藕花深处。\\
争渡，争渡，惊起一滩鸥鹭。
\end{tcolorbox}\\

\begin{tcolorbox}[skin=widget,
boxrule=1mm,
coltitle=black,
colframe=my-blue!45!white,
colback=my-blue!15!white,
width=(.9\linewidth),
before=\hfill,
after=\hfill,
center,]
\centering
\textbf{Prompt(translated into English)}\\
\justifying
Please write a poem in the form “Rumengling”. The theme or instruction is “cheerful”. Please strictly follow the number of character requirements for each line. Here is an example:
\begin{quote}
\centering
I often recall the sun setting at the riverside pavilion,
lost in the intoxication and unaware of the way back.\\
Later on when my excitement wanes, I return on the boat, only to find myself unwittingly entering a lotus pond.\\
Struggling to cross, struggling to cross, with seagulls and herons startled by me and flew away.
\end{quote}
\end{tcolorbox}\\

\newpage
\section{Relationship between Format Accuracy and Poem Length.}

We performed a regression analysis to investigate how format accuracy changes with poem length. Results show that in general the format accuracy decreases as the poem length increases. Results also show that CharPoet is less sensitive to poem length compared to competing models.

\begin{figure}[h]
    \centering    \includegraphics[width=0.47\textwidth]{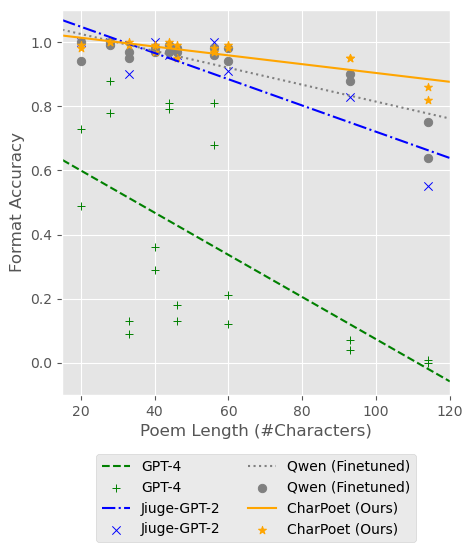}
    \caption {Relationship between Format Accuracy and Poem Length. Regression analysis indicates that hte format accuracy of CharPoet is less sensitive to increase in the poem length.}
    \label{fig:format_accuracy_regression}
\end{figure}

\end{CJK*}
\end{document}